\documentclass{article}
\usepackage[utf8]{inputenc}
\usepackage[T1]{fontenc}

\usepackage{times}
\usepackage{amsfonts} 
\usepackage{amssymb,amsmath,color}
\usepackage{natbib}
\usepackage{graphicx}
\usepackage{subfig}
\usepackage{float}

\newcommand{\eq}[1]{Eq.~(\ref{eq:#1})}
\newcommand{\mysec}[1]{Section~\ref{sec:#1}}

\newcommand{\BEAS}{\begin{eqnarray*}}
\newcommand{\EEAS}{\end{eqnarray*}}
\newcommand{\BEA}{\begin{eqnarray}}
\newcommand{\EEA}{\end{eqnarray}}
\newcommand{\BEQ}{\begin{equation}}
\newcommand{\EEQ}{\end{equation}}
\newcommand{\BIT}{\begin{itemize}}
\newcommand{\EIT}{\end{itemize}}
\newcommand{\BNUM}{\begin{enumerate}}
\newcommand{\ENUM}{\end{enumerate}}
\newcommand{\BA}{\begin{array}}
\newcommand{\EA}{\end{array}}

\newcommand{\Diag}{\mathop{\rm Diag}}

\newcommand{\tr}{\mathop{ \rm tr}}

\newcommand{\idm}{I}
\newcommand{\rb}{\mathbb{R}}
\newcommand{\N}{\mathbb{N}}

\newtheorem{proposition}{Proposition}

\def \X{{\mathcal X}}
\def \E{{\mathbb E}}
\def \1{{\mathbf 1}}
\def \P{{\mathbb P}}

\usepackage{fancyhdr}
\usepackage[normalem]{ulem}

\usepackage{authblk}

\title{Learning  Determinantal Point Processes in Sublinear Time}

\author[1,2]{Christophe Dupuy}
\author[1,3]{Francis Bach}
\date{}
\affil[1]{INRIA - Sierra project - team}
\affil[2]{Technicolor R\&D France}
\affil[3]{D\'epartement Informatique de l'\'Ecole Normale Sup\'erieure Paris}

\date{}

\begin{document}
\maketitle

\begin{abstract}
We propose a new class of determinantal point processes (DPPs) which can be manipulated for inference and parameter learning in potentially sublinear time in the number of items. This class, based on a specific low-rank factorization of the marginal kernel, is particularly suited to a subclass of continuous DPPs  and DPPs defined on exponentially many items. We apply this new class to modelling 
text documents as sampling a DPP of sentences, and propose a conditional maximum likelihood formulation to model topic proportions, which is made possible with no approximation for our class of DPPs. We present an application to document summarization with a DPP on $2^{500}$ items.
\end{abstract}

\section{Introduction}
Determinantal point processes (DPPs) show lots of promises for modelling diversity in combinatorial problems, e.g., in recommender systems or text processing \citep{kulesza2011k,gillenwater2012discovering,gillenwater2014expectation}, with algorithms for sampling \citep{kang2013fast,affandi2013approximate,li2016efficient,li2016fast} and likelihood computations based on linear algebra \citep{mariet2015fixed,gartrell2016low,kulesza2012determinantal}.

While most of these algorithms have polynomial-time complexity, determinantal point processes are too slow in practice for large number $N$ of items to choose a subset from. Simplest algorithms have cubic running-time complexity and do not scale well to more than ${N=1000}$.  Some progress has been made recently to reach quadratic or linear time complexity in $N$ when imposing low-rank constraints, for both learning and inference  \citep{kronDPP,gartrell2016low}.

This is not enough, in particular for applications in continuous DPPs where the base set is infinite, and for modelling documents as a subset of all possible sentences: the number of sentences, even taken with a bag-of-word assumption, scales exponentially with the vocabulary size. Our goal in this paper is to design a class of DPPs which can be manipulated (for inference and parameter learning) in potentially sublinear time in the number of items $N$.

In order to circumvent even linear-time complexity, we consider a novel class of DPPs which relies on a particular low-rank decomposition of the associated positive definite matrices. This corresponds to an embedding of the $N$ potential items in a Euclidean space of dimension $V$.
In order to allow efficient inference and learning, it turns out that a single operation on this embedding is needed, namely the computation of a second-order moment matrix, which would take time (at least) proportional to $N$ if done naively, but may be available in closed form in several situations. This computational trick makes  a striking parallel with positive definite kernel methods \citep{scholkopf2001learning,shawe2004kernel}, which use the ``kernel trick'' to work in very high dimension at the cost of computations in a smaller dimension.

In this paper we make the following contributions:
\BIT
\item[--] We propose in \mysec{DPP} a new class of determinantal point processes (DPPs) which is based on a particular low-rank factorization of the marginal kernel. Through the availability of a particular second-moment matrix, the complexity for inference and learning tasks is polynomial in the rank of the factorization and thus often sublinear in the total number of items (with exact likelihood computations).  

\item[--] As shown in \mysec{examples}, these new DPPs are particularly suited to a subclass of continuous DPPs (infinite number of items), such as on $[0,1]^m$, and DPPs defined on the $V$-dimensional hypercube, which has $2^V$ elements.

\item[--] We propose in \mysec{summary} a model of documents as sampling a DPP of sentences, and propose a conditional maximum likelihood formulation to model topic proportions. We present an application to document summarization with a DPP on $2^{500}$ items.

\EIT

\section{Review of Determinantal Point Processes}

In this work, for simplicity, we consider a very large \emph{finite} set $\X$, with cardinality $|\X| = N$, following~\citet{kulesza2012determinantal}.
In several places, we will consider an infinite set (see, e.g., \mysec{infinite}) \citep{affandi2013approximate,lavancier2015determinantal}. 

A determinantal point process (DPP) on a set $\X$ is a probability measure on $2^{\X}$, the set of all subsets of $\X$.
It can either be represented by a $L$-ensemble $L(x,y)$, for $x,y \in \X$ or by its marginal kernel $K(x,y)$, which we refer to as the ``$K$-representation'' and the ``$L$-representation''.
In this paper, $K$ and $L$ will be $N \times N$ matrices, with elements $K(x,y)$ and $L(x,y)$ for $x,y \in \X$.
Both $L$ and $K$ are potentially large matrices, as they are indexed by elements of $\X$.

A sample $X$ drawn from a DPP on $\X$ is a subset of $\X$, $X\subseteq\X$. 
In the ``$K$-representation'' of a DPP, for any set $A \subset \X$, we have:
\[
\P( A \subseteq X ) = \det K_A,
\]
where $K_A$ is the matrix of size $|A| \times |A|$ composed of pairwise evaluations of $K(x,y)$ for $x,y \in A$. If we denote by ``$\preccurlyeq$'' the positive semidefinite order on symmetric matrices (i.e., $A\preccurlyeq B \Leftrightarrow (B-A)$ is positive semidefinite), the constraint on $K$ is $ 0 \preccurlyeq K \preccurlyeq \idm$ so that the DPP is a probability measure. 

In the ``$L$-representation'', for any set $A \subset \X$, we have
\begin{equation}
\P(  X = A ) = \frac{ \det L_A}{\det ( \idm + L )}.
\label{eq:LLL}
\end{equation}
The constraint on $L$ is $L \succcurlyeq 0$.  

Given a DPP and its two representations $L$ and $K$, we can go from $L$ to $K$ as ${K = \idm - ( \idm  + L )^{-1}}$ and vice-versa as ${L = K ( \idm - K)^{-1}}$. The $L$-representation only exists when $K \prec\idm$, where ``$\prec$'' denotes the positive definite order of symmetric matrices (i.e., $A\prec B \Leftrightarrow (B-A)$ is positive definite).

Several tasks can be solved, e.g., marginalization, conditioning, etc., that are either easy in the $L$-representation or in the $K$-representation.
For instance, (conditional) maximum likelihood when observing sets is easier in the $L$-representation, as the likelihood of an observed set $A\subseteq\X$ is directly obtained with $L$ through \eq{LLL}. Conversely, the expected number of selected items, $\E[|X|]$ for a DPP defined by $L,K$ is easily computed with $K$ as ${\E[|X|]=\tr K}$ \citep{kulesza2012determinantal}.

The DPPs model aversion between items. For instance, if $X$ is drawn from a DPP($K,L$), the probability that items $i$ and $j$ are together included in $X$ is 
\[
\P(\{i,j\}\subseteq X) = K_{ii}K_{jj} - (K_{ij})^2.
\]
This probability then decreases with similarity $K_{ij}$ between item $i$ and item $j$. This key aversion property makes DPPs useful to document summarization (our application in \mysec{summary}) where we want to select sentences that covers the most the document while avoiding the selection of two similar sentences.

\paragraph{Approximate computations.}
In practice, the key difficulty is to deal with the cubic complexity in $|\X|$ of the main operations --- determinant and computations of inverses.
In their work, \cite{kulesza2012determinantal} propose a low-rank model for the DPP matrix $L$, namely ${L(x,y) = q(x)\langle \phi(x),\phi(y)\rangle q(y)}$, where ${q(x)\in\rb^+}$ corresponds to a ``quality'' measure of $x$ and $\phi(x)\in\rb^r$, ${\Vert\phi(x)\Vert=1}$ corresponds to the ``diversity'' feature (or embedding) of $x$. In matrix notations, we have ${L=\Diag(q)\Phi\Phi^\top\Diag(q)}$. In particular, they show that most of the computations are based on the matrix ${C=\Phi^\top\Diag(q^2)\Phi\in\rb^{r\times r}}$. As $\Phi\in\rb^{N\times r}$, they achieve an overall complexity $O(Nr^2)$. In their application to document summarization, they only parameterize and learn the ``quality'' vector $q$, fixing the diversity features $\Phi$.

More recently, \cite{gartrell2016low} use a low rank factorization of $L$ ($L=UU^\top$, with $U\in\rb^{N\times r}$) and apply accelerated stochastic gradient ascent on the log-likelihood of observed sets for learning $U$. They achieve a linear complexity in $N$: $O(Nr^2)$. \cite{kronDPP} propose a Kronecker factorization of $L$: ${L=L_1\otimes L_2}$ where $\otimes$ is the Kronecker product, ${L_i\in\rb^{N_i\times N_i}}$ and ${N=N_1N_2}$. They use a fixed point mehtod (with the Picard iteration) to maximize the likelihood, that consists in alternatively updating $L_1$ and $L_2$ with a computational complexity $O(N^{3/2})$ if ${N_1\approx N_2\approx \sqrt{N}}$.

However, when the set $\X$ is very large (e.g., exponential) or infinite, even linear operations in $N=|\X|$ are intractable. In the next sections, we provide a representation of the matrices $L$ and $K$ together with an optimization scheme that makes the optimization of the likelihood tractable even when the set $\X$ is too large to perform linear operations in $N=|\X|$.

\section{A Tractable Family of Kernels}
\label{sec:DPP}
We consider the family of matrices decomposed as a sum of the identity matrix plus a low-rank term, where the column-space of the low-rank term is fixed.
We show that if $K$ is in the family (with its additional constraint that $K \prec \idm$), so is $L$, and vice-versa.

\subsection{Low-rank family}
\label{sec:lowrank}
We consider a feature map $\phi: \X \to \rb^V$, a probability mass function $p: \X \to \rb_+$, a scalar $\sigma \in \rb_+$ and a symmetric matrix $B \in \rb^{V \times V}$. The kernel $K(x,y)$ is constrained to be of the form:  
\BEQ
\label{eq:K}
K(x,y) = \sigma 1_{x=y} + p(x)^{1/2}\phi(x)^\top B \phi(y) p(y)^{1/2} .
\EEQ
In matrix notation, this corresponds to   
$$K =  \sigma \idm + \Diag(p)^{1/2} \Phi B \Phi^\top \Diag(p)^{1/2}, $$
with $\Phi \in \rb^{|\X| \times V}$ and $B \in \rb^{V \times V}$. With additional constraints detailed below, this defines a valid DPP, for which the $L$-representation can be easily derived in the form
\BEQ
\label{eq:L}
L(x,y) = \alpha 1_{x=y} + p(x)^{1/2} \phi(x)^\top A \phi(y) p(y)^{1/2},
\EEQ
with $\alpha \in \rb_+$  and $A \in \rb^{V \times V}$.
The following proposition is a direct consequence of the  Woodbury matrix identity:
\begin{proposition}
The kernel $K$ defined in \eq{K} is a valid DPP if 
\\ \hspace*{.2cm}
(a)~${\sigma \in [0,1]}$,
\\ \hspace*{.2cm}
 (b)~ ${0 \preccurlyeq B \preccurlyeq (1-\sigma) ( \Phi^\top \Diag(p) \Phi)^{-1}}$. 
 
It corresponds to the matrix $L$ defined in \eq{L} with 
${\sigma= \frac{\alpha}{\alpha+1}}$ and \[B   =  \frac{1}{(\alpha+1)^2} \big[ A^{-1} + \frac{1}{\alpha+1} \Phi^\top \Diag(p)  \Phi \big] ^{-1}.\] 
\end{proposition}

Moreover, we may use the matrix determinant lemma to obtain
\BEAS
\det (L + \idm) & = & \det [ (\alpha+1) \idm ] \det [A] \det \Big( A^{-1} + \frac{1}{\alpha+1} \Phi^\top \Diag(p) \Phi \Big),
\EEAS
which is expressed through the determinant of a $V \times V$ matrix (instead of $N \times N$).

We may also go from $L$ to $K$ as ${\alpha=\frac{\sigma}{1-\sigma}}$ and 
${
A   =  \frac{1}{(1-\sigma)^2} \big[ B^{-1} - \frac{1}{1-\sigma} \Phi^\top \Diag(p)  \Phi \big] ^{-1}. 
}$

\subsection{Tractability}
From the identities above, we see that a sufficient condition for being able to perform the computation of $L$ and its determinant is the availability of the 
matrix 
$$\Sigma = \Phi^\top \Diag(p)  \Phi = \sum_{x \in \X} p(x) \phi(x) \phi(x)^\top \in \rb^{V \times V}.$$
In the general case, computing such an expectation would be (at least) linear time in $N$, but throughout the paper, we assume this is available in polynomial time in $V$ (and not in $N$). See examples in \mysec{examples}.

Note that this resembles the kernel trick, as we are able to work implicitly in a Euclidean space of dimension $N$ while paying a cost proportional to $V$. In our document modelling example, we will have $N = 2^V$, and hence we achieve sublinear time.

We can compute other statistics of the DPP when $K,L$ belong to the presented family of matrices. For instance, the expected size of a set $X$ drawn from the DPP represented by $K,L$ is: 
\BEQ
\label{eq:card}
\E[|X|] = \tr K =  \sigma |\X| + \tr \left(  B \Sigma \right).
\EEQ

Given $\phi(x)$, the parameters are the distribution $p(x)$ on $\X$, $A \in \rb^{V \times V}$ and $\alpha \in \rb_+$.
If $\X$ is very large, it is hard to learn $p(x)$ from observations and $p(x)$ is thus assumed fixed. We also have to assume that $\alpha$ is proportional to $1/|\X|$ or zero when $\X$ is infinite as the first term of $\E[|X|]$ in \eq{card}, $\frac{\alpha}{\alpha+1} |\X|$, must be finite.

\subsection{Additional low-rank approximation}

If $V$ is large, we use a low-rank representation for $A$: 
\begin{equation}
A = \gamma\idm +  U \Diag(\theta) U^\top,
\label{eq:A}
\end{equation}
 with $\gamma \in \rb_+$, $\theta \in \rb_+^r$ and $U \in \rb^{V \times r}$.  All the exact operations on $L$ are then linear in  $V$, i.e.,  as $O(V r^2)$. See details in Appendix~\ref{app:sents}. Moreover, the parameter $\theta$ can either be global or different for each observation, which gives flexibility to the model in the case where observations come from different but related DPPs on~$\X$. For instance, in a corpus of documents, the distance between words (conveyed by $U$) may be different from one document to another (e.g., \textit{field} and \textit{goal} may be close in a sport context, not necessary in other contexts). This can be modelled through $\theta$ as topic proportions of a given document (see \mysec{summary}).
 
Note that the additional low-rank assumption (\ref{eq:A}) corresponds to an embedding ${x\in\X\mapsto U^\top \phi(x)\in\rb^r}$, where $\phi(x)$ is fixed and $U$ is learned. 

\subsection{Infinite $\X$}
\label{sec:infinite}
Although we avoid dealing rigorously with continuous-state DPPs in this paper \citep{affandi2013approximate,affandi2014learning,bardenet2015inference}, we note that when dealing with exponentially large finite sets $\X$ or infinite sets, we need to set $\sigma=\alpha=0$ to avoid infinite (or too large) expectations for the numbers of sampled elements (which we use in experiments for ${N=2^V}$).

Note moreover, that in this situation, we have the kernel ${K(x,y) =  p(x)^{1/2} p(y)^{1/2} \phi(x)^\top B \phi(y)}$ and thus the rank of the matrix $K$ is at most $V$, which implies that the number of sampled elements has to be less than~$V$.

We can sample from the very large DPP as soon as we can sample from a distribution on $\X$ with density proportional to $p(x) \phi(x)^\top A \phi(x)$. Indeed, one can sample from a DPP by first selecting the eigenvectors of $L$, each with probability $\lambda_i/(\lambda_i+1)$---where the $\lambda_i$s are the eigenvalues of $L$---and then projecting the canonical basis vectors---one per item---on this subset of eigenvectors. The density for selecting the first item is proportional to the squared norm of the latter projection (see Algorithm~1 of \cite{kulesza2012determinantal} for more details). Given our fomula for~$L$, all the required densities can be expressed as being proportional to $p(x) \phi(x)^\top A \phi(x)$.  In our simulations, we use instead a discretized scheme.

\subsection{Learning parameters with maximum likelihood}
In this section, we present how to learn the parameters of the model, corresponding to the matrix~$A$.

We have access to the likelihood through observations.
We denote by $X_1,\dots,X_M$, with $X_i\subseteq\X$, the observations drawn from a density $\mu(X)$. Each set $X_i$ is a set of elements ${X_i=\{x_1^i,\ldots,x_{|X_i|}^i\}}$, with $x^i_j\in\X$. We denote by $\ell(X|L)$ the log-likelihood of a set $X$ given a DPP matrix $L$. Our goal is to maximize the expected log-likelihood under $\mu$, i.e., ${\E_{\mu(X)}[\ell(X|L)]}$. As we only have access to $\mu$ through observations, we maximize an estimation of ${\E_{\mu(X)}[\ell(X|L)]}$, i.e.,
${\mathcal{L}(L) = \frac{1}{M}\sum_{i=1}^M \ell(X_i|L)}$. As the log-likelihood of a set $X$ is ${\ell(X|L)=\log\det L_X -\log\det(L+\idm)}$, our objective function becomes:
\begin{equation}
\mathcal{L}(L)=\frac{1}{M}\sum_{i=1}^M\left(\log\det L_{X_i} - \log \det ( \idm + L )\right).
\label{eq:LL}
\end{equation}
In the following, we assume $p$ fixed and we only learn~$A$ in its form \eqref{eq:A}.

In practice we minimize a penalized objective, that is, for our parameterization of $A$ in \eq{A}, 
\[F(L)= -\mathcal{L}(L)+\lambda \mathcal{R}(U,\theta),\] where $\mathcal{L}$ is the log-likelihood of a train set of observations [\eq{LL}] and $\mathcal{R}$ is a penalty function. We choose the penalty ${\mathcal{R}(U,\theta)=\Vert\theta\Vert_1 +\Vert U\Vert_{1,2}^2}$ where $\Vert .\Vert_1$ is the $\ell^1$ norm and ${\Vert U\Vert_{1,2}=\sum_{i=1}^r \Vert u_i\Vert_2}$, where ${\Vert .\Vert_2}$ is the $\ell^2$ norm and ${u_i}$ is the $i^{th}$ column of~$U$. The group sparsity norm ${\Vert .\Vert_{1,2}^2}$ allows to set columns of $U$ to zero and thus learn the number of columns.

This is a non non-convex problem made non smooth by the group norm. Following \cite{lewis2013nonsmooth}, we use BFGS to reach a local optimum of our objective function.
\section{Examples}
\label{sec:examples}

In this section, we review our three main motivating examples: (a) orthonormal basis based expansions applicable to continuous space DPPs; (b) standard orthonomal embedding with ${\X=\{1,\ldots,N\}}$ and (c)~exponential set ${\X = \{0,1\}^V}$ for applications to document modelling based on sentences in \mysec{summary}.

\subsection{Orthonormal basis based expansions}
\label{sec:cont}

We consider a fixed probability distribution $p(x)$ on~$\X$ and an orthonormal basis of the Euclidean space of square integrable (with respect to $p$) functions on $\X$. We consider $\phi(x)_i$ as the value at $x$ of the $i$-th basis function. Note that this extends to any $\X$, even not finite by going to Hilbert spaces.

We consider $A = \Diag(a)$ and $B = \Diag(b)$ two diagonal matrices in $\rb^{V \times V}$. Since $\phi(x)$ is an orthonormal basis, we have:
$$
\Sigma = \sum_{x \in \X} p(x) \phi(x) \phi(x)^\top = \idm,
$$
with a similar result for any subsampling of $\phi(x)$ (that is keeping a subset of the basis vectors).

For example, for $\X = [0,1]$, $p(x)$ the uniform distribution, ${\alpha=0}$ and $\phi(x)$ the cosine/sine basis, we obtain the matrix ${L(x,y) = \phi(x)^\top \Diag(a) \phi(y)}$ which is a 1-periodic function of $x-y$, and we can thus model any of these functions.
This extends to $\X = [0,1]^m$ by tensor products, and hyperspheres by using spherical harmonics~\citep{atkinson2012spherical}.

\paragraph{Truncated Fourier basis.}
In practice, we consider the truncated Fourier orthonormal basis of $\rb^V$ with ${V=2d+1}$, that is, 
${\phi_1(x)=1}$, ${\phi_{2i}(x)= \sqrt{2} \cos ( 2\pi i x)}$
and ${\phi_{2i+1}=\sqrt{2}  \sin ( 2\pi i x)}$, for ${i \in \{1,\dots, d\}}$ and ${x\in\X}$. If ${A = \Diag(a)}$ is diagonal, then ${L(x,y) = \phi(x)^\top\Diag(a)\phi(y)}$ is a $1$-periodic function of $x-y$, with only the first $d$ frequencies, which allows us to learn covariance functions which are invariant by translation in the cube.
We could also use the $K$-representation ${K(x,y) = \phi(x)^\top B\phi(y)}$, with $B=\Diag(b)$ diagonal in $[0,1]$, but the log-likelihood maximization is easier in the $L$-representation.

We use this truncated basis to optimize the log-likelihood $\mathcal{L}(L)$ on finite observations [\eq{LL}] , i.e., ${X_i\subseteq\X}$ and ${|X_i|<\infty}$. In particular, the normalization constant is computed efficiently with this representation of $L$ as we have ${\det(L+\idm) = \prod_{i=1}^V (a_i+1)}$.

\paragraph{Non-parametric estimation of the stationary function.}
We may learn any 1-periodic function of $x-y$ for $L(x,y)$ or $K(x,y)$ and we do so by choosing the truncated Fourier basis of size~$V$, we could also use positive definite kernel techniques to perform non-parametric estimation.

\paragraph{Running time complexity.} For general continuous ground set ${\X =[0,1]^m}$, with ${m\geq 1}$, the running time complexity is still controlled by $V=(2d+1)^m$, $d$ corresponding to the number of selected frequencies in each dimension of the Fourier basis (with ${a\in\rb^{V}}$). The value of $d$ may be adjusted to fit the complexity in $O(V\kappa^3)$ or $O(d^m\kappa^3)$, where $\kappa$ is the size of the biggest observation (i.e., the largest cardinality of all observed sets).

\subsection{Standard orthonormal embeddings}
\label{sec:itemset}
In this section, we consider DPPs on the set ${\X=\{1,\ldots,V\}}$ (i.e., ${N=V}$). We choose the standard orthonormal embedding, that is ${\Phi=\idm}$ which gives ${L=\alpha\idm + U\Diag(\theta)U^\top}$, taking ${\gamma=0}$. For this particular model, the complete embedding ${\Phi U=U}$ is learned and the distribution $p(x)$ is included in $U$. This is only possible when $V$ is small. This model is suited to item selection, where groups of items are observed (e.g., shopping baskets) and we want to learn underlying embeddings of these items (through parameter~$U$). Again, the size of the catalog $V$ may be very large. Note that unlike existing methods leveraging low-rank representations of DPPs \citep{kronDPP,gartrell2016low}, the parameter $\theta$ in our representation can be different for each observation, which makes our model more flexible. 

\subsection{$\X = \{0,1\}^V$}
\label{sec:expset}
In this section, we consider DPPs on the set ${\X=\{0,1\}^V}$. For large values of $V$, direct operations on matrices $L,K$ may be impossible as $\X$ is exponential, ${|\X|=2^V}$. 
In particular, we consider the model where $\phi(x) = x$, i.e., ${\Phi\in\rb^{2^V\times V}}$ (in other words we simply embed~${\{0,1\}^V}$ in $\rb^V$).

As mentioned above, the tractability of the DPP($L,K$) on ${\X=\{0,1\}^V}$ depends on the expectation ${\Sigma=\sum_{x\in\X}p(x)xx^\top}$. For particular distributions $p(x)$, $\Sigma$ is computed in closed form. For instance, if $p(x)$ corresponds to $V$ independent Bernouillis, i.e., ${p(x) =  \prod_{i=1}^V \pi_i^{x_i} (1-\pi_i)^{1-x_i}}$, the expectation quantity is ${\Sigma=\Diag(\pi(1-\pi)) +  \pi \pi^\top}$. If the independent Bernoullis are exchangeable, i.e., all $\pi$'s are equal, we have ${\Sigma=\pi(1-\pi) \idm + \pi^2  1 1^\top }$.

The tractability of our model is extended to the case $\X = \N^V$ with Poisson variables. Indeed, if ${p(x)=\prod_{i=1}^V \frac{e^{-\lambda_i} \lambda_i^{x_i}}{x_i!}}$ for $x\in\N^V$, the expectation of~${xx^\top}$ over $\X$ is ${\Sigma=\Diag(\lambda)+\lambda\lambda^\top}$. 

Given these examples, we assume in the rest of the paper that:
\[
\Sigma=\sum_{x\in\X}p(x)xx^\top = \Diag(\nu)+\mu\mu^\top.
\]

The complexity of operations on the matrix $L$ with this structure is $O(Vr^2)$ --- instead of $O( 2^Vr^2)$ if working directly with $L$; see Appendix~\ref{app:sents} for details. If we use the factorization of $\Sigma$ above:
\BEAS
\tr K & = & \frac{\alpha}{\alpha+1} 2^V  \\
&& + \frac{1}{(\alpha+1)^2} \tr \bigg[  \big( \Diag(\nu)+  \mu\mu^\top \big)\\
&& \times  \Big( \frac{1}{\alpha+1}   ( \Diag(\nu)+  \mu\mu^\top ) + A^{-1} \Big)^{-1} \bigg].
\EEAS
This identity suggests that we replace $\alpha$ by $\alpha 2^{-V}$ in order to select a finite set $X\subseteq\X$. For large values of~$V$, $\alpha=0$ is the key choice to avoid infinite number of selected items.

\section{DPP for document summarization}
\label{sec:summary}
We apply our DPP model to document summarization.
Each document $X$ is represented by its sentences, ${X=(x_1,\ldots,x_{|X|})}$ with $x_i \in\X=\{0,1\}^{V}$. The variable $V$ represents the size of the vocabulary, i.e., the number of possible words. A sentence is then represented by the set of words it contains, ignoring their exact count and the order of the words.
We want to extract the summary of each document as a subset of observed sentences. We use the structure described in \mysec{lowrank} to build a generative model of documents.
Let $K\in\rb^{2^V \times 2^V}$ be the marginal kernel of a DPP on the possible sentences $\X$.
We consider that the summary $Y\subseteq\X$ of document $X$ is generated from the DPP($K,L$) as follows:
\begin{enumerate}
\item Draw sentences $X=(x_1,\ldots,x_{|X|})$ from DPP represented by $L$,
\item Draw summary $Y\subseteq X$ from DPP represented by~$L_X$.
\end{enumerate}
In practice, we observe a set of documents and we want to infer the word embeddings $U$ and the topic proportions $\theta$ for each document. In the following we consider that $\alpha$ and $\gamma$ are fixed. We also denote by $\mathcal{L}(U,\theta)\equiv\mathcal{L}(L)$ the log-likelihood of observations [\eq{LL}] for simplicity as our DPP matrix $L$ is encoded by $U$ and $\theta$.

The intuition behind this generative model is that the sentences of a document cover a particular topic (the topic proportions are conveyed by the variable $\theta$) and it is very unlikely to find sentences that have the same meaning in the same document. In this sense, we want to model aversion between sentences of a document.

\paragraph{Parameter learning.}
As explained in \mysec{expset}, we assume ${\Sigma = \Diag(\nu) + \mu\mu^\top}$. The log-likelihood of an observed document $X$ is ${\ell(X|L)=\log\det L_X - \log\det(L+\idm)}$. The computation of the second term, ${\log\det(L+\idm)}$, is untractable to compute in reasonable time for any $L$ when $V\geq 20$, since ${L\in\rb^{2^V\times 2^V}}$. We can still compute this value exactly for structured $L$ coming from our model with complexity $O(Vr^2)$ (see Appendix~\ref{app:sents} for details).

We infer the parameters $U$ and $\theta$ by optimizing our objective function ${F(U,\theta)= -\mathcal{L}(U,\theta)+\lambda \mathcal{R}(U,\theta)}$ with respect to $U$ and $\theta$ alternatively. In practice, we perform 100 iterations of L-BFGS for the function ${U\mapsto F(U,\theta)}$ and 100 iterations of L-BFGS for each function $\theta_i\mapsto F(U,\theta)$, for~$i=1,\ldots,M$. The optimization in $U$ can also be done with stochastic gradient descent (SGD) \citep{OnlineLearning1998}, using a mini-batch $D_t$ of observations at iteration $t$: ${U\leftarrow U - \rho_t G_{D_t}(U)}$, with ${G_{D_t}(U)}$ the unbiased gradient ${G_{D_t}(U)= -\frac{1}{|D_t|}\sum_{i\in D_t}\nabla_U\ell(X_i|L(U,\theta_i)+\lambda\nabla_U\mathcal{R}(U,\theta)}$. We choose L-BFGS over SGD for settings simplicity. In particular, the choice of the stepsize $\rho$ and the mini-batch size $|D_t|$ is not straightforward.

\section{Experiments}
\subsection{Datasets}
We run experiments on synthetic datasets generated from the different types of DPPs described above. For all the datasets, we generate the observations using the sampling method described by \citet{kulesza2012determinantal} (Algorithm~1 page 16) and perform the evaluation for 10 different datasets. This method draws exact samples from a DPP matrix $L$ and its eigendecomposition (which requires $N$ to be less than 1000). For the evaluation figures, the mean and the variance over the 10 datasets are respectively displayed as a line and a shaded area around the mean.

\begin{figure}[t]
\centering
\subfloat[DPP.]{\includegraphics[width=0.35\textwidth]{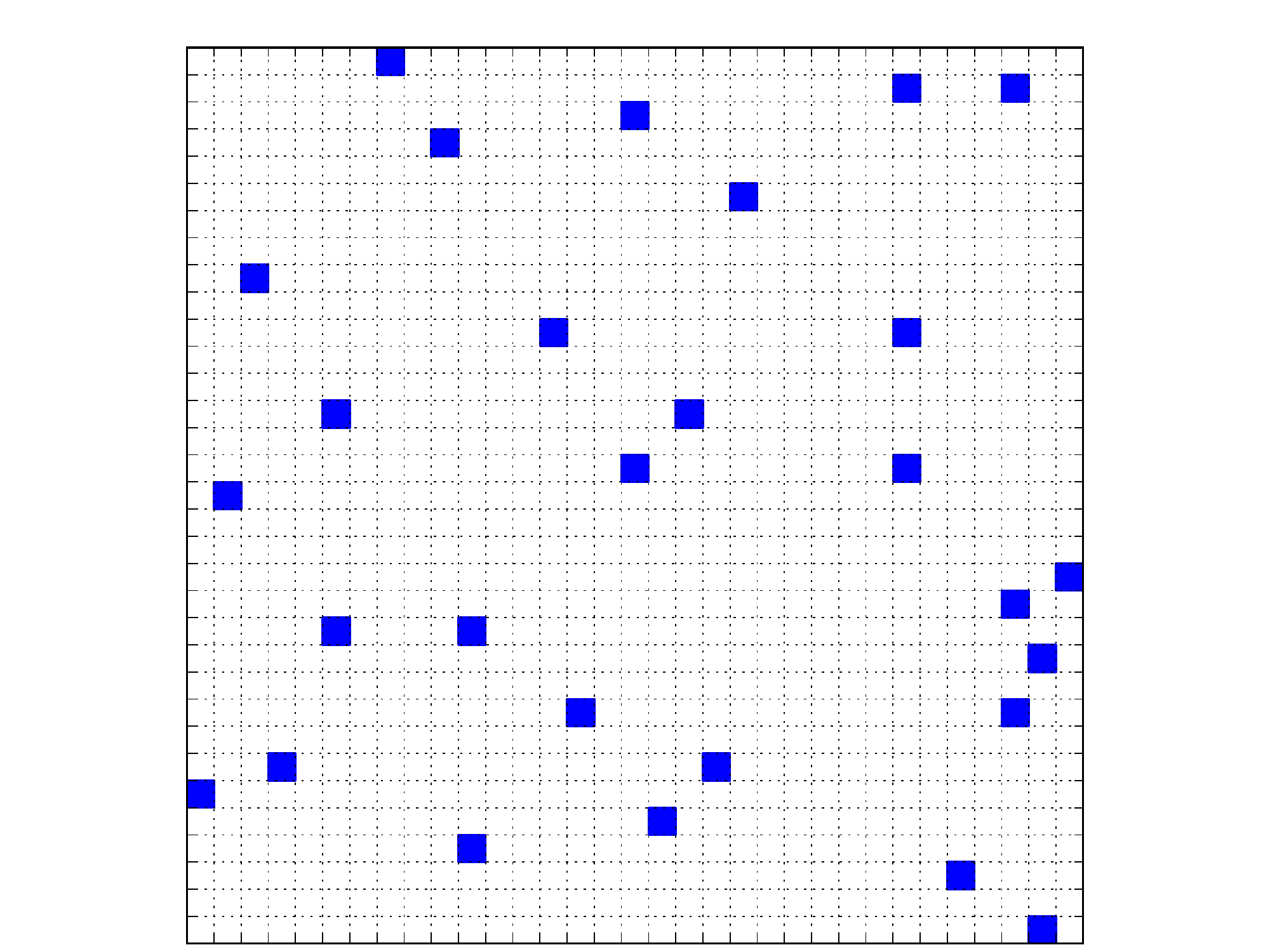}}
\subfloat[I.i.d.]{\includegraphics[width=0.35\textwidth]{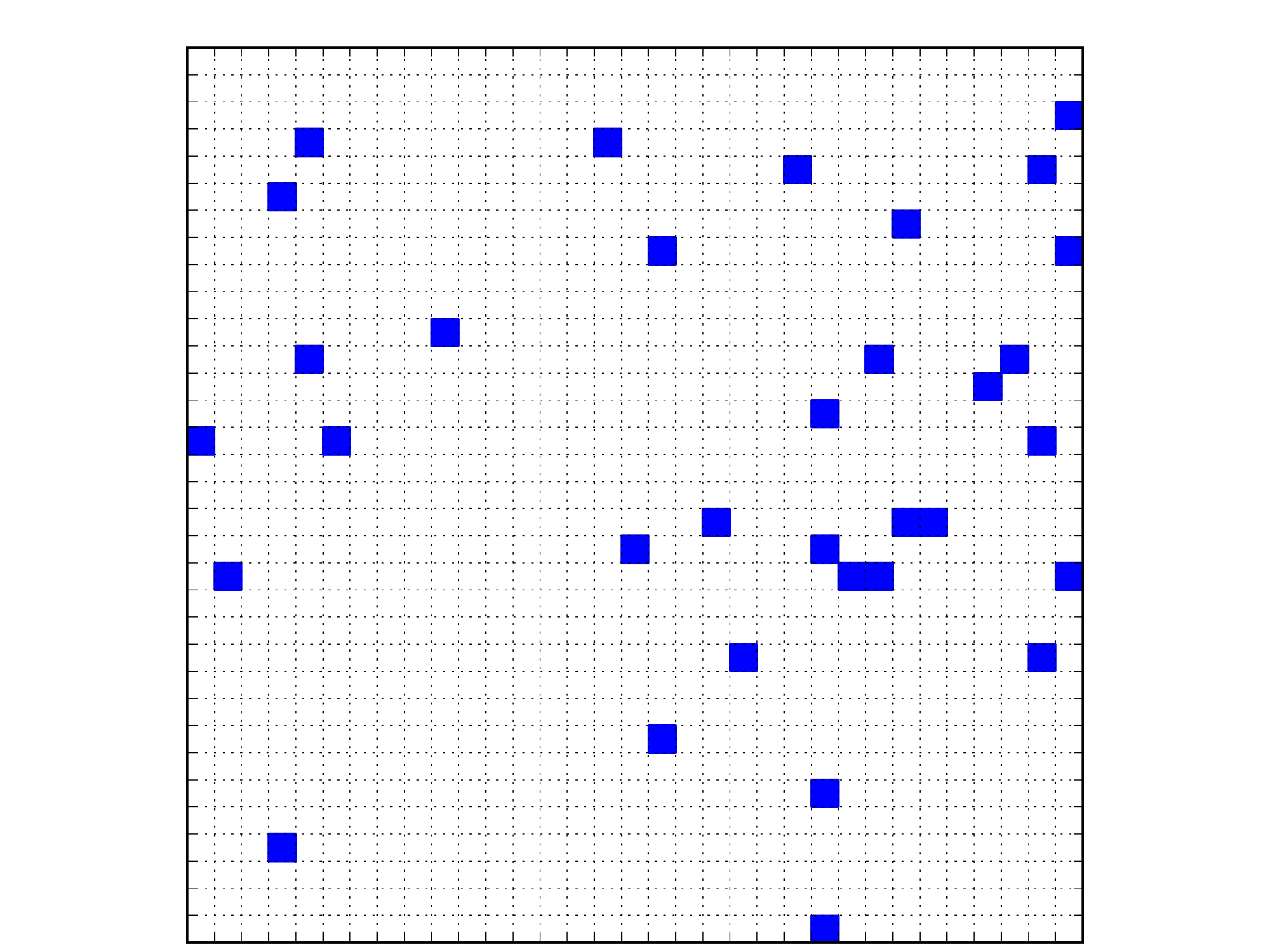}}
\vspace*{-0.1cm}
\caption{Comparison of points drawn from a DPP (left) independently from uniform distribution (right).}
\label{fig:sample}
\end{figure}

\paragraph{Continuous set $[0,1]^m$.}
We describe in \mysec{cont} a method to learn from subsets drawn from a DPP on a continuous set $\X$. As sampling from continuous DPPs is not straightforward and approximate \citep{affandi2013approximate}, we consider a discretization of the set $[0,1]^2$ into the discrete set ${\{0,1/N,\dots,(N-1)/N\}^2}$. Note that this discretization only affects the sampling scheme.
We generate a dataset from the ground set ${\X = \{0,1/N,\dots,(N-1)/N\}^2}$ with the DPP represented by ${L(x,y) = \phi(x)^\top\Diag(a)\phi(y)}$ , with embedding ${\phi(x)\in\rb^{N^2}}$ the discrete Fourier basis of ${(\rb^{N})^2}$, i.e., for ${(i,j)\in\{1,\ldots,N\}^2}$, ${\phi(x)_{i,j}=\psi(x_1)_i\psi(x_2)_j}$ with ${\psi(z)_{1}=1}$, ${\psi(z)_{2k}=\sqrt{2}\cos(2\pi kz)}$ and ${\psi(z)_{2k+1}=\sqrt{2}\sin(2\pi kz)}$, for ${k=1,\ldots,(N-1)/2}$. With notations of \mysec{cont} we have $V=N^2$. For ${(i,j)\in\{0,\ldots,N-1\}}^2$, we set ${a_{(i,j)}=C_iC_j\tilde{a}_i\tilde{a}_j}$, with $C_0=1$, ${\tilde{a}_0=1}$ and ${C_i=1/\sqrt{2}}$, ${\tilde{a}_i=1/i^\beta}$ for~${i\geq 1}$. We choose ${N=33}$ (i.e., ${V=N^2=1069}$) and ${\beta=2}$ for the experiments.
We present in Figure~\ref{fig:sample} two samples: a sample drawn from the DPP described above and a set of points that are i.i.d. samples from the uniform distribution on $\X$. We observe aversion between points of the DPP sample that are distributed more uniformly than points of the i.i.d. sample.

\paragraph{Items set.}
We generate observations from the ground set $\X=\{1,\ldots,V\}$, which corresponds to the matrix ${L=\alpha\idm+U\Diag(\theta)U^\top}$. For these observations, we set ${V=100}$, ${r=5}$, $\alpha=10^{-5}$. For each dataset, we generate $U$ and $\theta$ randomly with different seeds accross the datasets.

\paragraph{Exponential set.}
We generate observations from the ground set $\X=\{0,1\}^V$ with ${\phi(x)=x}$. In this case, we set ${V=10}$, ${r=2}$, $\alpha=10^{-5}$, $\gamma=1/V$. As we need the eigendecomposition of $L\in\rb^{2^V\times 2^V}$, we could not generate exact samples with higher orders of magnitude for $V$. However, we can still optimize the likelihood for ground sets with large values of $V$ and we run experiments on real document datasets, where the size of the vocabulary is~${V=500}$ (i.e., $\vert\X\vert=2^{500}\approx 10^{150}$).

For both ground sets ${\X=\{1,\ldots V\}}$ and ${\X=\{0,1\}^N}$, we consider two types of datasets: one dataset where all the observations are generated with the same DPP matrix $L$ and another dataset where observations are generated with a different matrix $L(\theta^i)$ for each observation. For the second type of dataset, the embedding~$U$ is common to all the observations while the variable~$\theta^i$ differs from one observation to another.

\begin{figure}[t]
\centering
\includegraphics[width=0.4\textwidth]{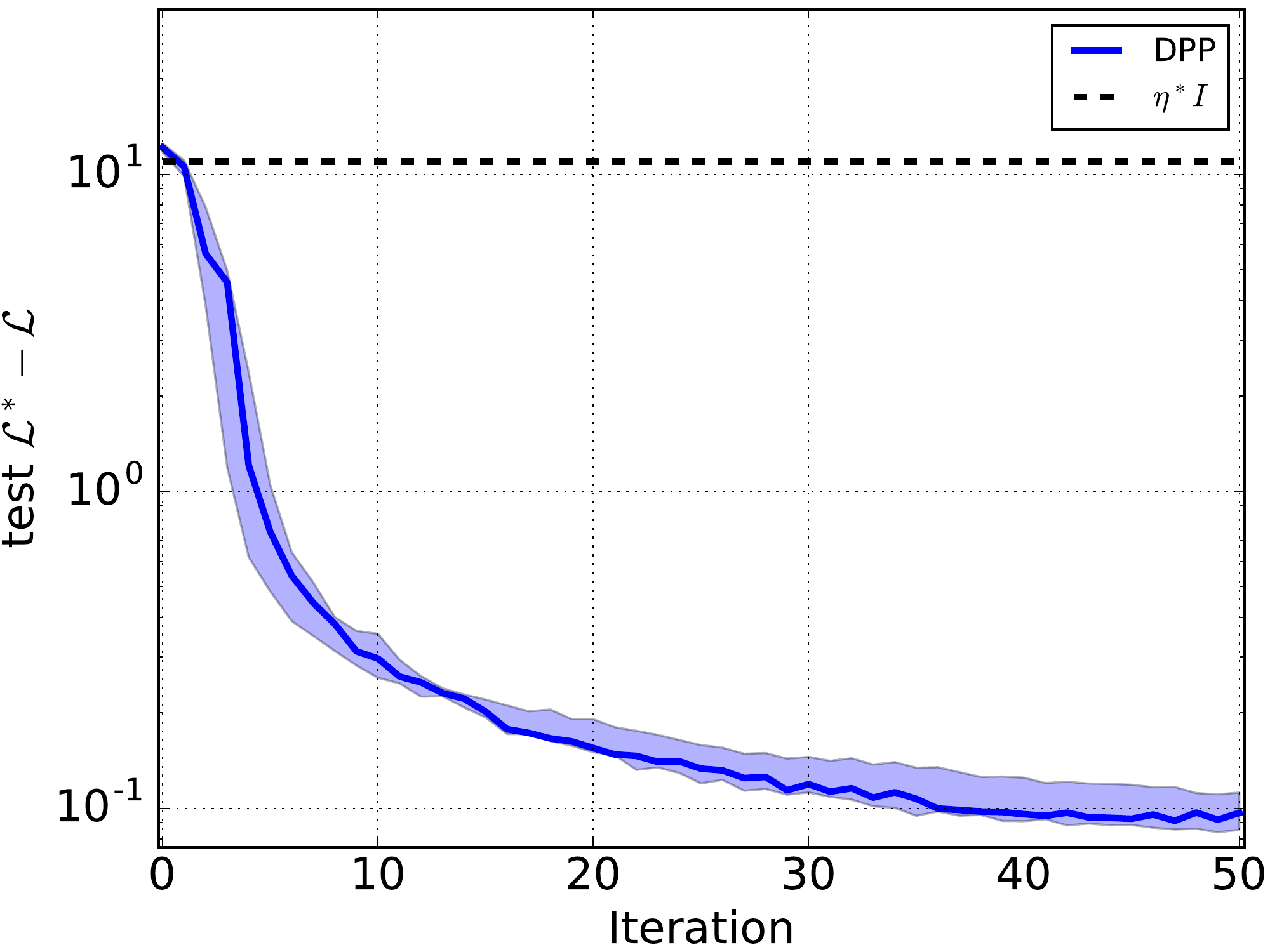}
\vspace*{-0.1cm}
\caption{Continuous set $[0,1]^2$. Distance in log-likelihood ($\mathcal{L}^*-\mathcal{L}$).}
\label{fig:fourier}
\end{figure}

\begin{figure*}[t]
\centering
\subfloat[Distance in log-likelihood.]{\includegraphics[width=0.3\textwidth]{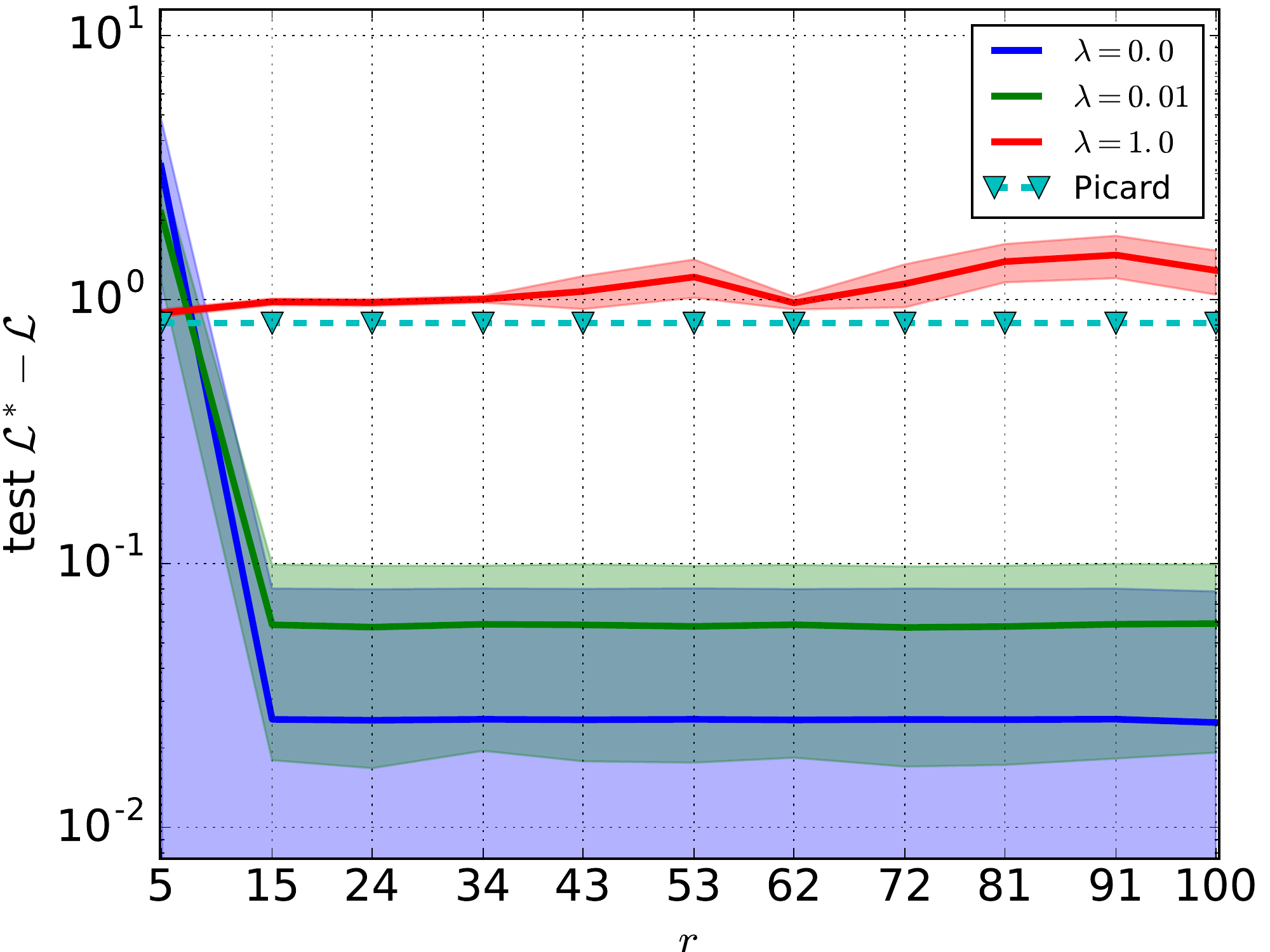}}
\subfloat[Distance between $U$ and $U^*$.]{\includegraphics[width=0.3\textwidth]{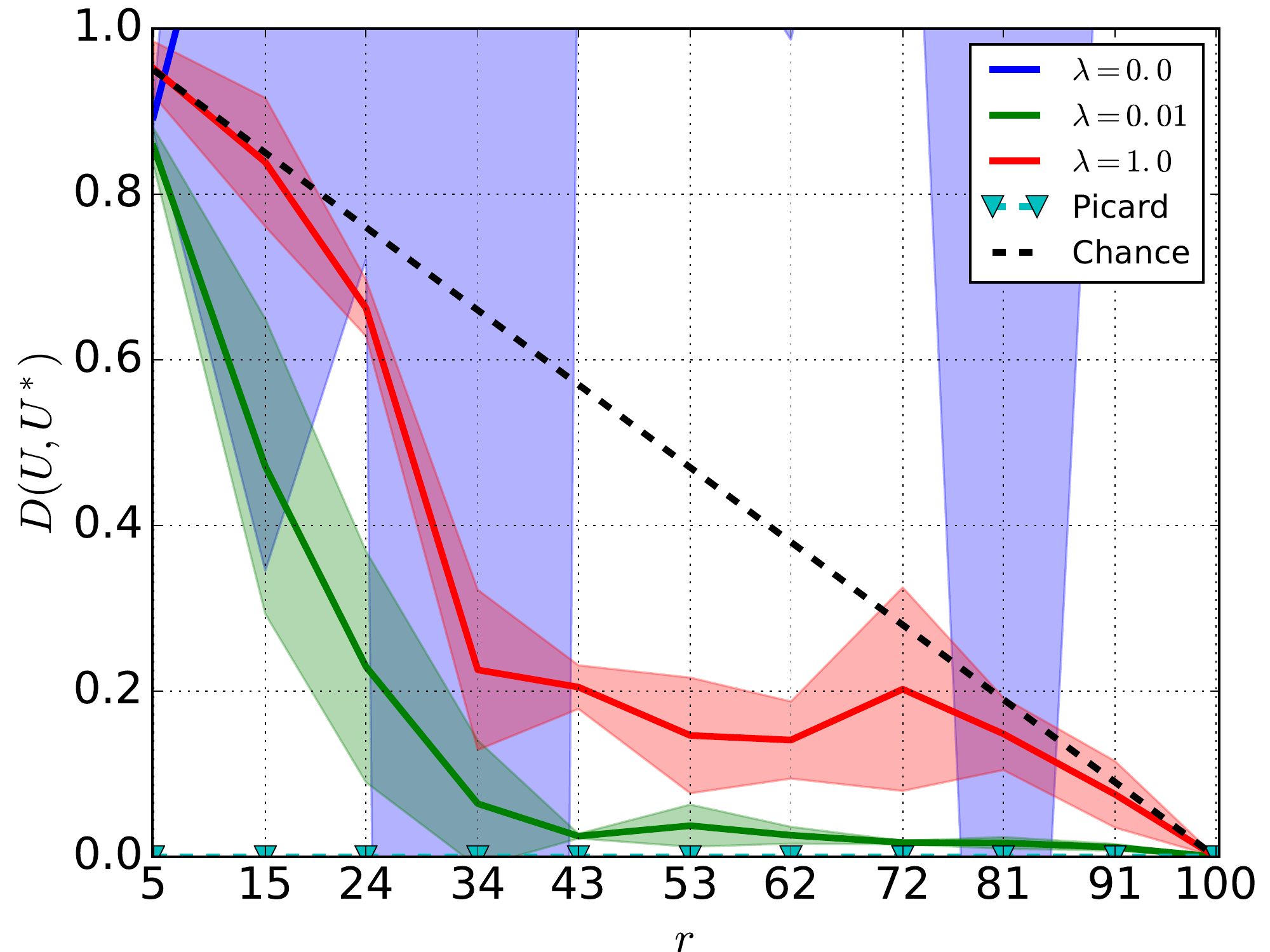}}
\subfloat[ Distance between $U$ and $U^*$.]{\includegraphics[width=0.3\textwidth]{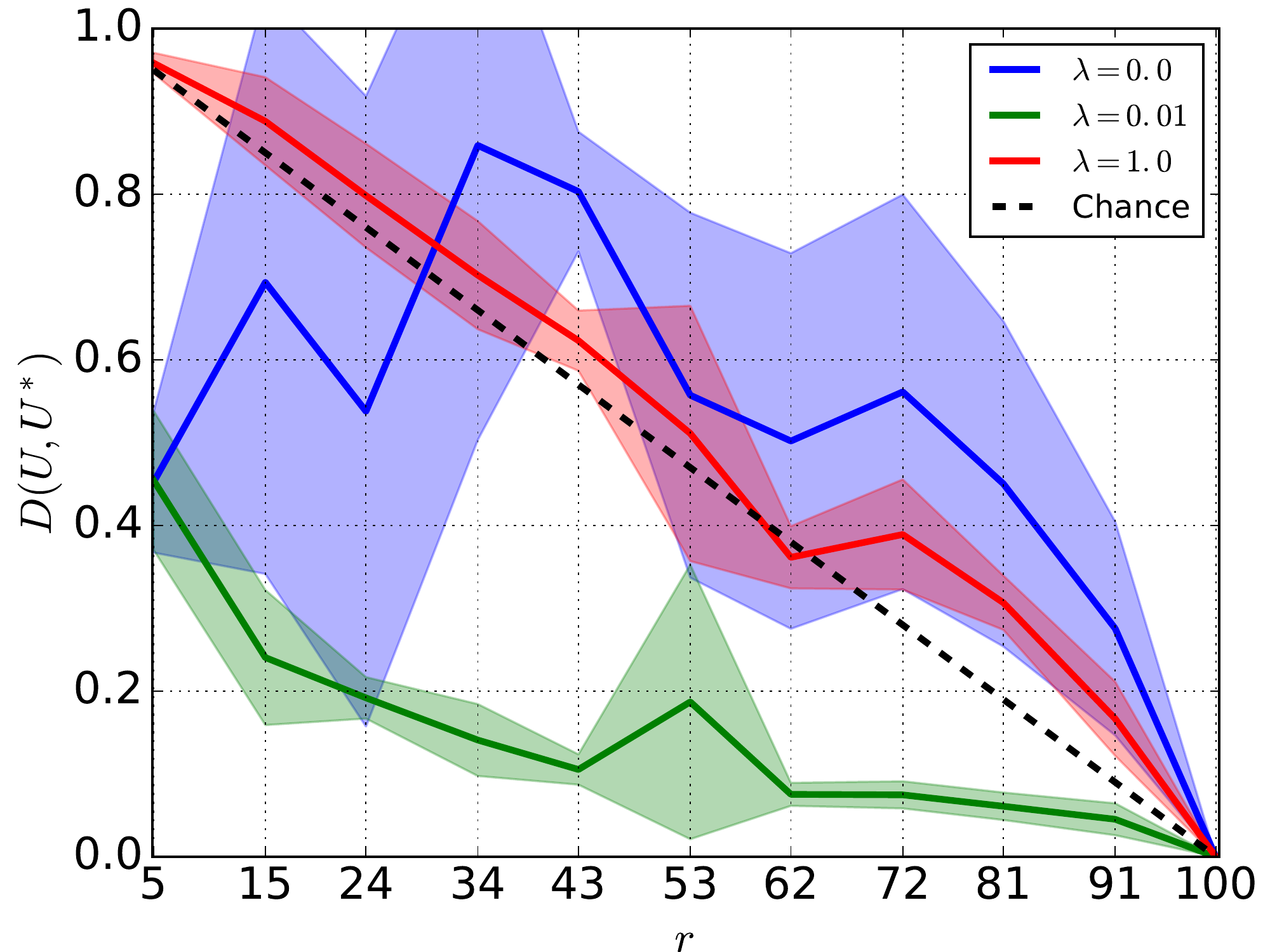}}
\vspace*{-0.1cm}
\caption{Performance for ground set ${\X=\{1,\ldots,V\}}$ as a function of $r$. (a,b)~Same $\theta$ for all the observations; (c)~A~different $\theta$ for each observation.}
\label{fig:items}
\end{figure*}

\begin{figure*}[t]
\centering
\subfloat[Distance in log-likelihood.]{\includegraphics[width=0.3\textwidth]{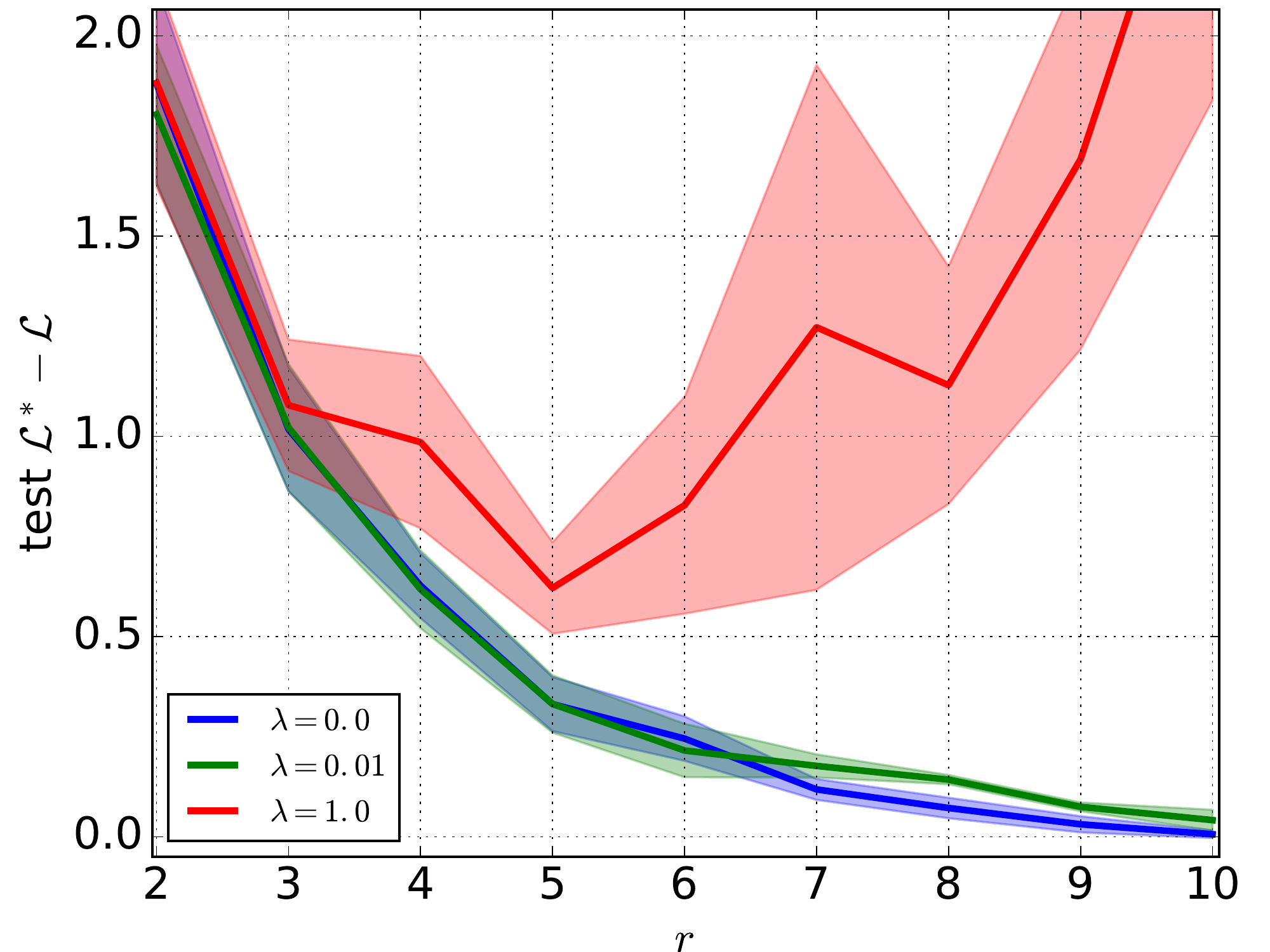}}
\subfloat[Distance between $U$ and $U^*$.]{\includegraphics[width=0.3\textwidth]{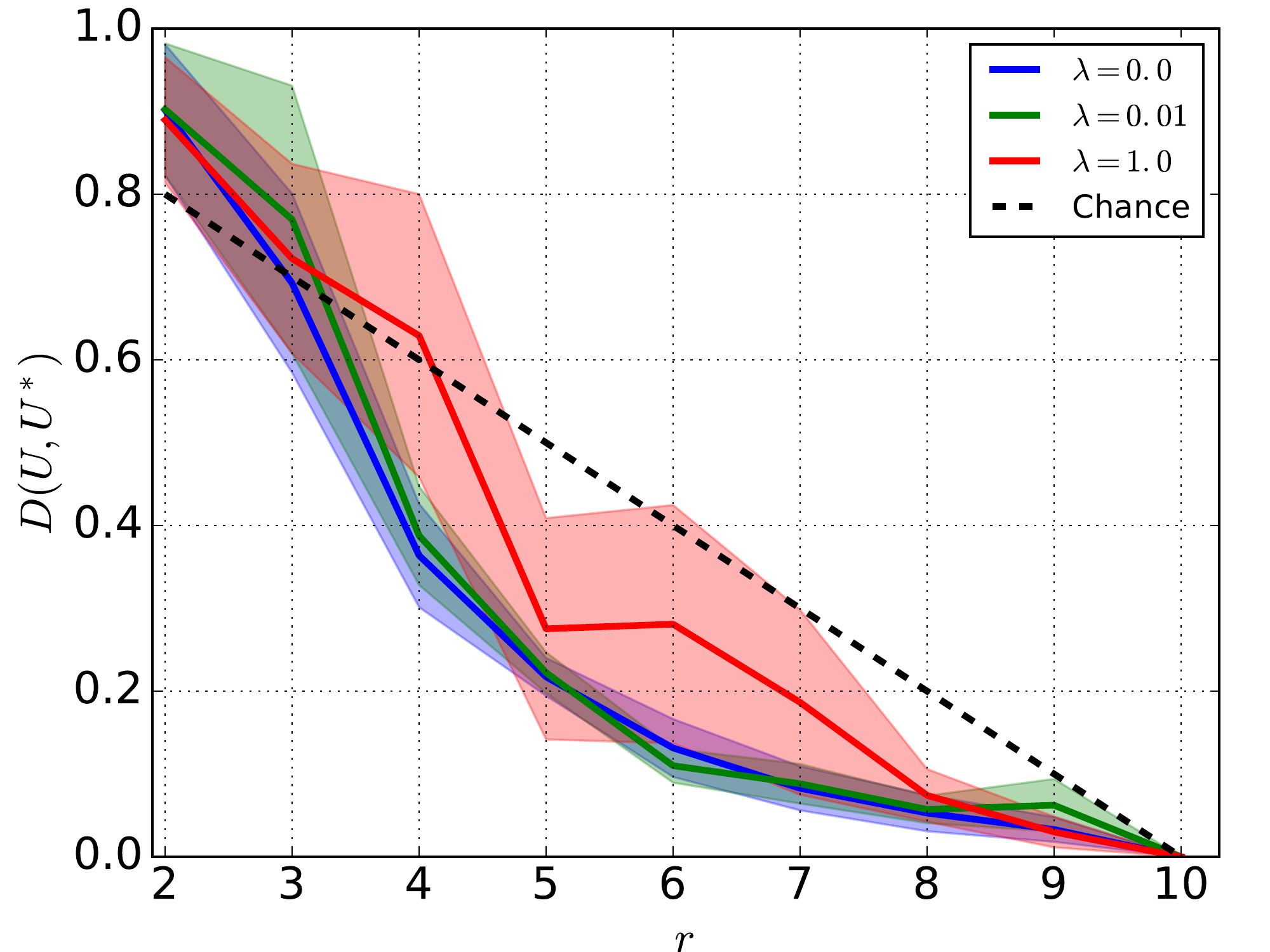} }
\subfloat[Distance between $U$ and $U^*$.]{\includegraphics[width=0.3\textwidth]{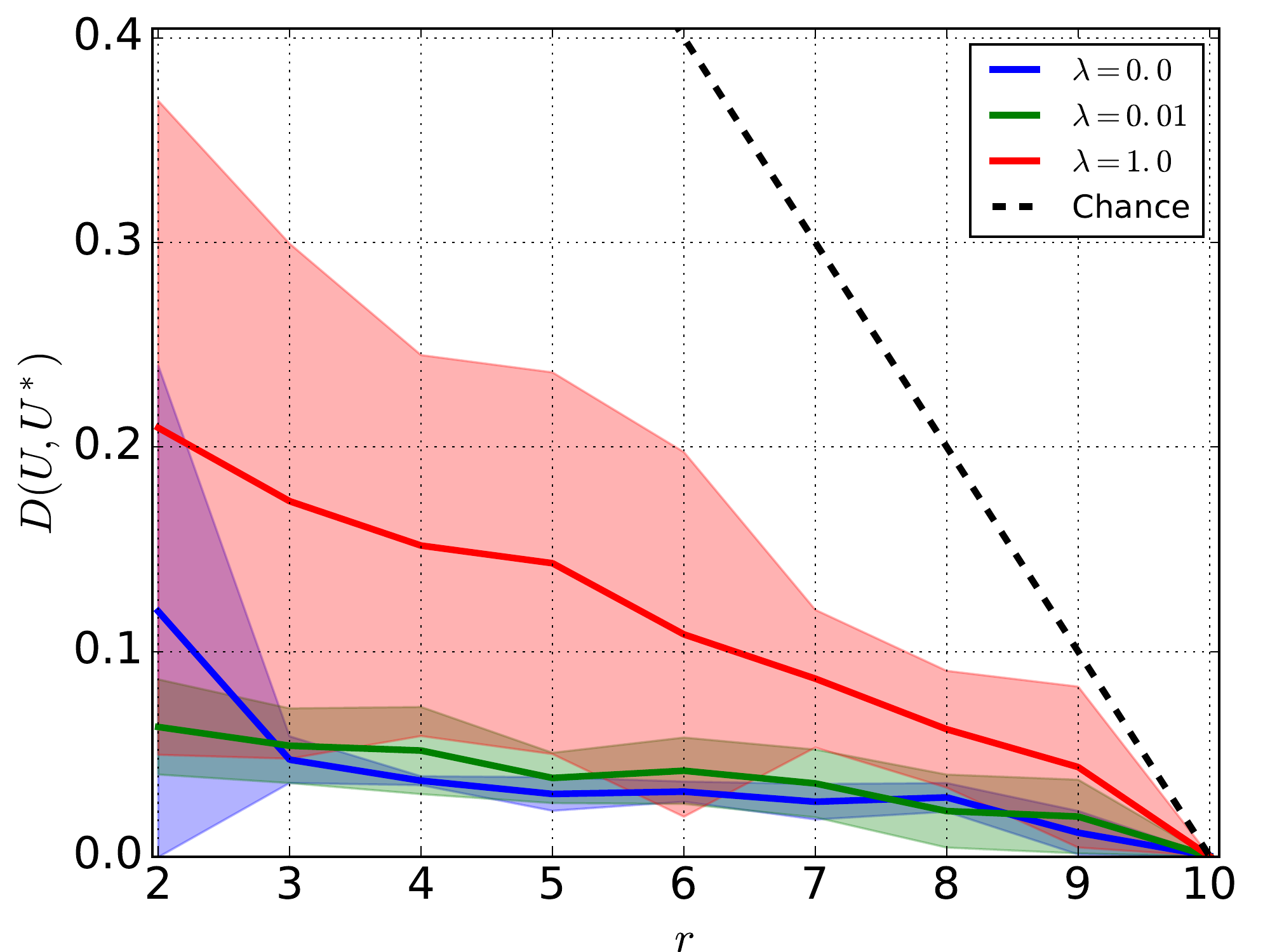}}
\vspace*{-0.1cm}
\caption{Performance for ground set ${\X=\{0,1\}^V}$ as a function of $r$. (a,b)~Same $\theta$ for all the observations; (c)~A~different $\theta$ for each observation.}
\label{fig:sents}
\end{figure*}

\begin{table*}[t]
\centering
\caption{Examples of reviews with extracted summaries (of size $l=5$ sentences) colored in blue.}
\vspace*{-0.4cm}
\label{tab:summary}
\begin{center}
\begin{tabular}{p{0.97\textwidth}}
\hline
{\bf Review 1}\\
\hline
Ate here once each for dinner and Sunday brunch.  {\bf \color{blue}{[Dinner was great.]}}  {\bf \color{blue}{[We got a good booth seat and had some tasty food.]}} I ordered just an entree since I wasn't too hungry.  The guys ordered appetizers and salad and I couldn't resist trying some. The risotto with rabbit meatballs was so good. {\bf \color{blue}{[Corn soup, good.]}}  {\bf \color{blue}{[And my duck breast, also good.]}} I was happy.  {\bf \color{blue}{[The sides were good too.]}} Potatoes and asparagus.   Came back for Mother's Day brunch.  Â Excellent booth table at the window, so we could watch our valeted car.  Pretty good service.  Good food.  No complaints. \\
\hline
\hline
{\bf Review 2}\\
\hline
This will be my 19 month old's first bar.  :D  I came here with a good friend and my little guy.  We shared the double pork chop and the Mac n Cheese. {\bf \color{blue}{[The double pork chop was delicious.....]}} {\bf \color{blue}{[Huge portions and beautifully prepared vegetables.]}} {\bf \color{blue}{[What a wonderful selection of butternut squash, spinach, cauliflower and mashed potato.]}}  We were very impressed with the chop, meat was tender and full of flavor.  {\bf \color{blue}{[The mac n cheese, was okay.]}}  I would definitely go back for the pork chop...   might want to try the fried mushrooms too.  {\bf \color{blue}{[Place surprisingly was pretty kid friendly.]}} The bathroom actually had a bench I could change my little guy!\\
\hline
\end{tabular}
\end{center}
\end{table*}

\paragraph{Real dataset.}
We consider a dataset of 100,000 restaurant reviews and minimize the objective function ${F(U,\theta)}$ mentioned above. We first remove the stopwords using the NLTK toolbox \citep{NLTK}. Among the remaining words, we only keep the $V=500$ most frequent words of the dataset. After filtering, the average number of sentences per review is 10.5 and each sentence contains on average 4.5 words. We use the proposed DPP structure to (1)~learn word embedding $U$ from observations and (2)~extract a summary for each review using the model of \mysec{summary}. Given a document $X$, the inferred parameters $U$ and $\theta(X)$ and the corresponding DPP matrix $L$, we extract the $l$ sentences summarizing the document $X$ by solving the following maximization:
\[
Y^* \in\arg\max_{Y\subseteq X,\; |Y|=l}\;\; \frac{\det (L_Y)}{\det(L_X+\idm)}.
\]
In practice we use the greedy MAP algorithm \citep{gillenwater2012near} to extract the summary $\widehat{Y}$ of document $X$, as an approximation of the MAP $Y^*$ (with the usual submodular maximization approximation guarantee \citep{krause2012submodular}).

\subsection{Evaluation}
We evaluate our optimization scheme with two metrics. First, we compare the log-likelihood on the test set obtained with the inferred model $\mathcal{L}$ to the test log-likelihood with the model that generated the data $\mathcal{L}^*$. We use this metric when the data is generated with a single set of parameters over the dataset (i.e., the same DPP matrix $L$ is used to generate all the observations) as in such case the difference of test log-likelihood between two models ($\mathcal{L}^*-\mathcal{L}$) is an estimation of the Kullback-Leibler divergence between the two models.

We also consider a distance between the inferred embedding $U$ and the embedding that generated the data~$U^*$. As the performance is invariant to any permutation of column in the matrix $U$ (together with indices of~$\theta$) and to a scaling factor --- both $(U,\theta)$ and $(\frac{1}{\sqrt{\gamma}}U,\gamma\theta)$ correspond to the same DPP matrix $L$ --- we consider the following distance that compares the linear space produced with ${U\in\rb^{V\times r}}$ and ${U^*\in\rb^{V\times r^*}}$:
\begin{align*}
D(U,U^*) & = \frac{\Vert U(U^\top U)^{-1}U^\top U^* - U^*\Vert_F}{\Vert U^*\Vert_F},
\end{align*}
where $\Vert .\Vert_F$ is the Frobenius norm. This distance is invariant to scaling and rotation and is equal to zero when $U$ and $U^*$ span the same space in $\rb^V$. In particular, if we generate randomly the $r$ columns of $Z\in\rb^{V\times r}$, the expectation of the distance to $U^*$ is ${\E_Z[D(Z,U^*)] = 1-\frac{r}{V}}$. We display this quantity as ``chance'' in the following.

\paragraph{Continuous set ${[0,1]^2}$.}
We compare our inference method to the best diagonal DPP ${L_{\eta^*}=\eta^*\idm}$, where ${\eta^*\in\rb}$ maximizes the log-likelihood. 

\paragraph{Items set, ${\X=\{1,\ldots,V\}}$.} We compare our inference method to the Picard iteration on full matrices proposed by \cite{mariet2015fixed}. As they only consider the scenario where all the observations are drawn from the same DPP, we only compare to our method in that case.

\subsection{Results}
\paragraph{Continuous set ${[0,1]^2}$.} We present the difference in log-likelihood between the inferred model and the model that generates the data as a function of the iterations in Figure~\ref{fig:fourier}. 
The comparison between the resulting kernel and the kernel that generates the data is presented in Appendix~\ref{app:fourier}. We observe that our model performs significantly better than the $\eta^*\idm$ kernel and converges to the the true log-likelihood.

\paragraph{Items set \& exponential set.}
We present the difference in log-likelihood and the distance of embeddings $U$ between the inferred model and the model that generates the data as a function of the rank~$r$ of the representation in Figure~\ref{fig:items} for the gound set~${\X=\{1,\ldots,V\}}$ and in Figure~\ref{fig:sents} for the ground set~${\X=\{0,1\}^V}$. We observe that 
the penalization may deteriorate the performance in terms of log-likelihood but significantly improves the quality of the recovered parameters. In practice, as our penalization~$\mathcal{R}$ induces sparsity we recover sparse $\theta$ when~${r>r^*}$. For both ground sets, the parameter $U^*$ that generated the data is recovered  for ${r^*\leq r<V}$.

For the items set ${\X=\{1,\ldots,V\}}$, while the datasets are generated with ${r^*=5}$, we observe that the parameter $U^*$ is only recovered when we optimize with $r\geq 30$.
We also observe that our method performs better than the Picard iteration of \cite{mariet2015fixed} in terms of log-likelihood. The Picard iteration updates the full matrix $L$ and there is no tradeoff between the rank and the closeness of spanned subspaces, conveyed by ${D(U,U^*)}$.

For the exponential set ${\X=\{0,1\}^V}$, ${r^*=2}$ and the parameter $U^*$ is recovered for $r\geq 6$.

\paragraph{Real dataset.}
Summaries with $l=5$ sentences of two reviews are presented in Table~\ref{tab:summary}. The corresponding embeddings $U$ are presented in Appendix~\ref{app:topics}. We observe that our method is able to extract sentences that describes the opinion of the user on the restaurant. In particular, the sentences extracted with our method convey commitment of the user to aspects (food, service,...) while other sentences of the reviews only describe the context of the meal.

\section{Conclusion}
\label{sec:ccl}

In this paper, we proposed a new class of determinantal point processes that can be run on a huge number of items because of a specific low-rank decomposition. This allowed parameter learning for continuous DPPs and new applications such as document modelling and summarization.

We apply our model on exponential set ${\X=\{0,1\}^V}$ to model documents, it would be interesting to apply our inference to the infinite ground set ${\X=\N^V}$ as suggested in the paper. We would also like to study the inference in continous exponential set ${\X=\rb^V}$ using our low-rank decomposition.

While we focused primarily on DPPs to model diversity, it would also be interesting to consider other approaches based on submodularity 
\citep{djolonga2014map,Djolonga16Mixed} and study the tractability of these models for exponantially large numbers of items.

\section*{Acknowledgements}
We would like to thank Patrick Perez for helpful discussions related to this work.

\bibliographystyle{plainnat}
\bibliography{bib}

\newpage
\onecolumn
\clearpage
\appendix

\section{Continuous set $[0,1]^2$}
\label{app:fourier}
\begin{figure}[h]
\centering
\includegraphics[width=0.8\textwidth]{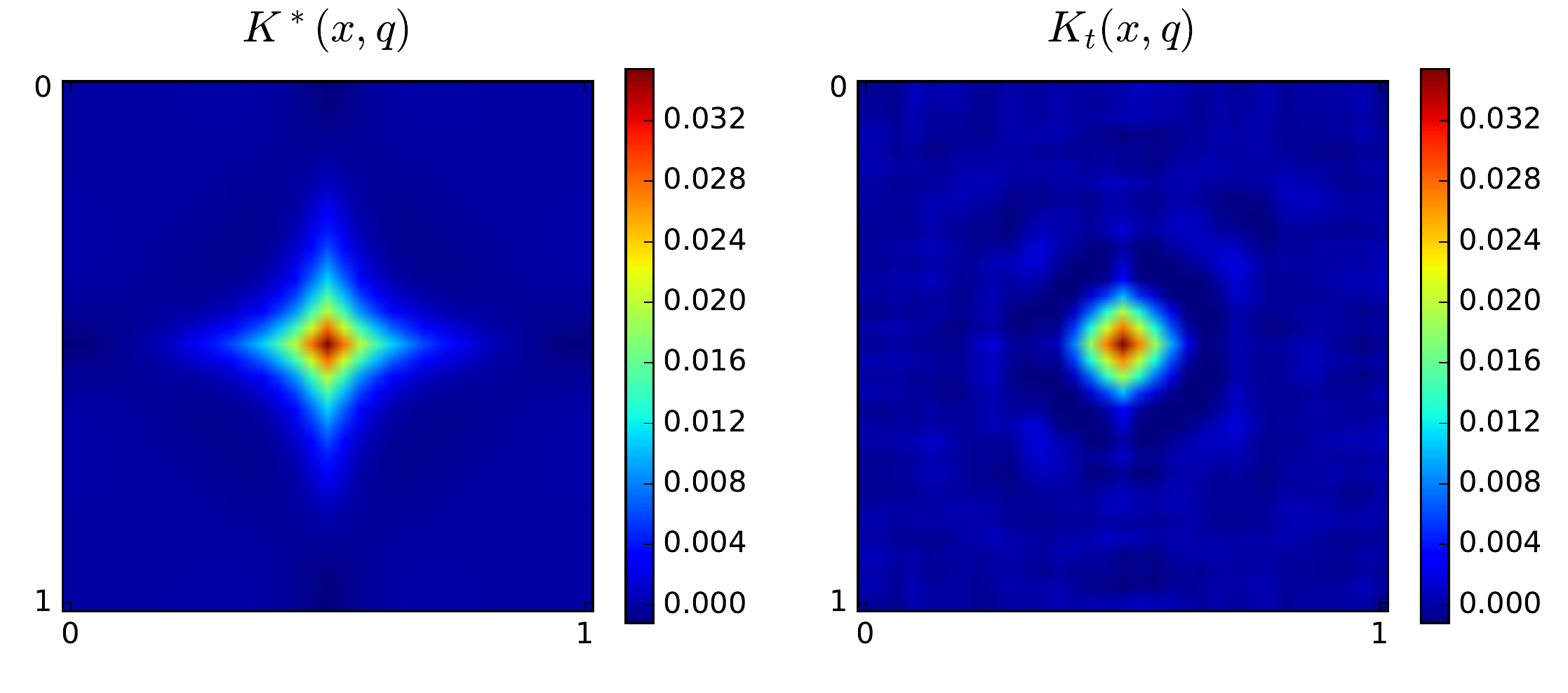}
\caption{Comparison of $K^*$ and $K_t$.}
\label{figapp:FourierK}
\end{figure}

In this section, we present a comparison between the true marginal kernel (that generates the data) $K^*$ and the inferred marginal kernel $K_t$. More precisely, ${\X=[0,1]^2}$ and we compute the induced distance from the center point ${q=(\frac{1}{2},\frac{1}{2})}$ to any point $x\in\X$, i.e., $K(x,q)$. We show in Figure~\ref{figapp:FourierK} a comparison between the true distance $K^*(x,q)$ and the inferred distance $K_t(x,q)$ after $t=100$ iterations.

\section{Picard iteration}
\begin{figure}[h]
\centering
\includegraphics[width=0.5\textwidth]{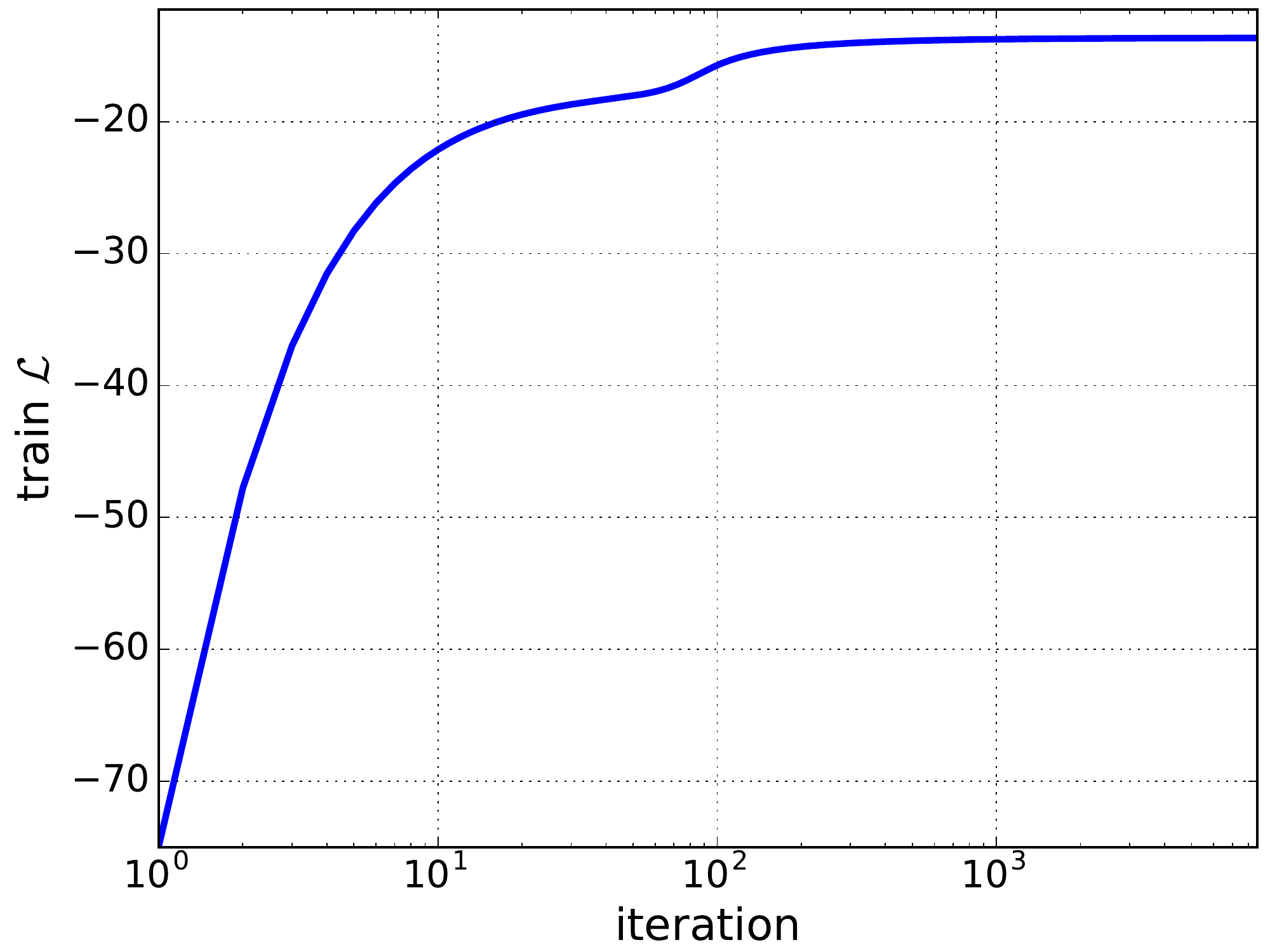}
\caption{Picard iteration \citep{mariet2015fixed}. Evolution of the objective function (train log-likelihood) as a function of the iterations.}
\label{figapp:picard}
\end{figure}

We apply the Picard iteration of \cite{mariet2015fixed} on the synthetic ``items'' datasets (i.e., observations are generated from ${L=\alpha\idm+U\Diag(\theta)U^\top}$) with $N=100$ items.
We present the evolution of the objective function through the iterations with the Picard iteration in Figure~\ref{figapp:picard}. We observe a similar evolution than presented in the original paper \citep{mariet2015fixed}. This however led in Figure~\ref{fig:items} to a lower likelihood than L-BFGS on~$U$.

\newpage
\section{Summary as a subsample -- Parameter learning}
\label{app:sents}
We assume that $\sum_{x \in \X} p(x) \phi(x)\phi(x)^\top = \Diag(\nu) + \mu\mu^\top$. The log-likelihood of an observed document $X$ is expressed as ${\ell(X|L)=\log\det L_X - \log\det(L+\idm)}$. The computation of the second term, ${\log\det(L+\idm)}$, is untractable to compute in reasonable time for any $L$ when $V\geq 20$, since ${L\in\rb^{2^V\times 2^V}}$. We can still compute this value for structured $L$. When ${L=\alpha\idm+\Diag(p)^{1/2}\Phi A\Phi^\top\Diag(p)^{1/2}}$, we have, using the matrix determinant lemma and Woodbury identity:
\[
\det (L + \idm) = \det [ (\alpha+1) \idm ] \det A \det \left( A^{-1} + \frac{1}{\alpha+1} \Phi^\top \Diag(p) \Phi \right).
\]

We then have, if $\rho = \frac{1}{\alpha+1}$:
\begin{align*}
\log\det\big(A^{-1} + \rho \Diag(\nu) +&  \rho \mu\mu^\top\big) = \log\left[1+\rho \mu^\top\left(A^{-1}+\rho\Diag(\nu)\right)^{-1}\mu\right]+\log\det(A^{-1}+\rho \Diag(\nu)) \\
& \text{ (matrix determinant lemma)}\\
 = &\log\left[1+\mu^\top\left(\Diag(1/\nu)-\Diag(1/\nu)(\rho A+\Diag(1/\nu))^{-1}\Diag(1/\nu)\right)\mu\right]\\
&+\log\det(A^{-1}+\rho\Diag(\nu))\\
&\text{(Woodbury identity)}.\\
\end{align*}
If we consider $A=\gamma I +U\Diag(\theta)U^\top$, we have:
\begin{align*}
(\rho A+\Diag(1/\nu))^{-1}  = &\left[(\rho\gamma I+ \rho U\Diag(\theta)U^\top +\Diag(1/\nu) \right]^{-1}\\
= &\Diag\left(\frac{\nu}{1+\nu\rho\gamma}\right)\\
- \Diag\left(\frac{\nu}{1+\nu\rho\gamma}\right)U &\left(\Diag(1/\rho\theta)+U^\top\Diag\left(\frac{\nu}{1+\nu\rho\gamma}\right) U\right)^{-1}U^T\Diag\left(\frac{\nu}{1+\nu\rho\gamma}\right),\\\\
\log\det(A^{-1}+\rho \Diag(\nu)) =&\log\det\left[\Diag(\frac{1}{\gamma}+\rho \nu)-\frac{1}{\gamma}U\left(\Diag(\gamma/\theta)+U^\top U\right)^{-1}U^\top\right]\\ & \text{ (Woodbury identity on }A^{-1})\\
 = &\log\det\left[\left(\Diag(\gamma/\theta)+U^\top U\right) -\frac{1}{\gamma}U^\top\Diag\left(\frac{\gamma}{1+\nu\gamma\rho}\right) U\right]\\
 & -\log\det(\Diag(\gamma/\theta)+U^\top U) +\log\det(\frac{I}{\gamma}+\rho\Diag(\nu))\\
= &\log\det\left[\Diag(\gamma/\theta)+U^\top\Diag\left(\frac{\nu\gamma\rho}{1+\nu\gamma\rho}\right) U\right]\\
&-\log\det(\Diag(\gamma/\theta)+U^\top U) +\log\det(\frac{I}{\gamma}+\rho\Diag(\nu)),\\\\
\log\det(A) = & \log\det(\Diag(1/\theta) + \frac{1}{\gamma}U^\top U) + \sum_k \log\theta_k + V\log\gamma
\end{align*}
In the end, the computation of $\log\det(L+I)$ only needs matrix products of size $V$ and inversions of size~$r$.

\newpage
\section{Results on real datasets}
\label{app:topics}
\subsection{Columns of $U$}
We present five embeddings (i.e., columns of $U\in\rb^{V\times r}$) out of $r=10$ learned on a restaurant reviews dataset with our DPP structure in Table~\ref{tab:topics} below. We display the 20 words with the highest absolute values for each column of $U$. We observe that our embeddings extract qualitative words (e.g., \textit{good}, \textit{great}, \textit{friendly}). Even if the embeddings are not as consistent as topics extracted with topic models (e.g., LDA), we can distinguish different aspects of restaurants with the embeddings. For instance, words with positive values in embedding 1 are related to the food (e.g., \textit{cream}, \textit{love}, \textit{crispy}, \textit{tomato}); words with positive values in embedding 2 are associated to the service aspect (with \textit{service}, \textit{friendly}, \textit{staff}, \textit{attentive}). Moreover, they already lead to good summaries.

\begin{table*}[h]
\caption{Five embeddings (columns of $U$) inferred with $r=10$ on restaurant reviews dataset.}
\label{tab:topics}
\vspace*{-0.5cm}
\begin{center}
\small
\begin{tabular}{|ll|}
\hline
{\bf Embed. 1} & {\bf $U_{w,1}$} \\
\hline
 love  &  0.19  \\
 could  &  0.11  \\
 large  &  0.11  \\
 cream  &  0.1  \\
 crispy  &  0.1  \\
 tomato  &  0.09  \\
 meat  &  0.09  \\
 ice  &  0.08  \\
 sauce  &  0.08  \\
 mouth  &  0.08  \\
\vdots & \vdots \\
 back  &  -0.48  \\
 sushi  &  -0.51  \\
 place  &  -0.51  \\
 pretty  &  -0.53  \\
 really  &  -0.58  \\
 come  &  -0.64  \\
 great  &  -0.73  \\
 service  &  -0.79  \\
 food  &  -1.08  \\
 good  &  -2.08  \\
\hline
\end{tabular}
\begin{tabular}{|ll|}
\hline
{\bf Embed. 2} & $U_{w,2}$ \\
\hline
 service  &  0.74  \\
 friendly  &  0.36  \\
 nice  &  0.33  \\
 good  &  0.24  \\
 pretty  &  0.24  \\
 staff  &  0.24  \\
 price  &  0.15  \\
 experience  &  0.14  \\
 well  &  0.14  \\
 attentive  &  0.13  \\
\vdots & \vdots \\
 would  &  -0.26  \\
 think  &  -0.27  \\
 try  &  -0.28  \\
 restaurant  &  -0.28  \\
 one  &  -0.3  \\
 amazing  &  -0.33  \\
 like  &  -0.57  \\
 get  &  -0.64  \\
 love  &  -0.73  \\
 place  &  -1.05  \\
\hline
\end{tabular}
\begin{tabular}{|ll|}
\hline
{\bf Embed. 3} & $U_{w,3}$ \\
\hline
 great  &  0.99  \\
 food  &  0.8  \\
 service  &  0.4  \\
 star  &  0.32  \\
 worth  &  0.26  \\
 place  &  0.26  \\
 price  &  0.25  \\
 back  &  0.21  \\
 wait  &  0.2  \\
 definitely  &  0.18  \\
\vdots & \vdots \\
 also  &  -0.27  \\
 tasty  &  -0.28  \\
 fresh  &  -0.31  \\
 salad  &  -0.32  \\
 delicious  &  -0.36  \\
 really  &  -0.4  \\
 nice  &  -0.43  \\
 like  &  -0.44  \\
 chicken  &  -0.45  \\
 order  &  -0.59  \\
\hline
\end{tabular}
\begin{tabular}{|ll|}
\hline
{\bf Embed. 4} & $U_{w,4}$ \\
\hline
 place  &  0.75  \\
 great  &  0.41  \\
 good  &  0.35  \\
 really  &  0.28  \\
 love  &  0.21  \\
 nice  &  0.16  \\
 service  &  0.16  \\
 atmosphere  &  0.14  \\
 get  &  0.14  \\
 friendly  &  0.13  \\
\vdots & \vdots \\
 could  &  -0.21  \\
 dinner  &  -0.21  \\
 menu  &  -0.25  \\
 restaurant  &  -0.27  \\
 well  &  -0.28  \\
 come  &  -0.37  \\
 eat  &  -0.38  \\
 food  &  -0.39  \\
 time  &  -0.4  \\
 price  &  -0.4  \\
\hline
\end{tabular}
\begin{tabular}{|ll|}
\hline
{\bf Embed. 5} & $U_{w,5}$ \\
\hline
 back  &  0.49  \\
 come  &  0.37  \\
 try  &  0.34  \\
 definitely  &  0.3  \\
 get  &  0.28  \\
 would  &  0.26  \\
 wait  &  0.23  \\
 dinner  &  0.15  \\
 friend  &  0.15  \\
 recommend  &  0.14  \\
\vdots & \vdots \\
 small  &  -0.17  \\
 atmosphere  &  -0.19  \\
 love  &  -0.21  \\
 restaurant  &  -0.22  \\
 everything  &  -0.24  \\
 say  &  -0.24  \\
 nice  &  -0.25  \\
 delicious  &  -0.26  \\
 great  &  -0.34  \\
 food  &  -0.53  \\
\hline
\end{tabular}
\end{center}
\end{table*}

\clearpage
\subsection{Rows of $U$}
From the embeddings $U$, we can also compute similarity between words using the rows of $U$. We use the cosine similarity, i.e., for words $v,w\in\{1,\ldots,V\}$:
\[
\mathrm{Cos}(v,w) = \frac{\langle U_v,U_w\rangle}{\Vert U_v\Vert_2 \Vert U_w\Vert_2},
\]
where $U_v\in\rb^r$ is the $v^{th}$ row of $U$.
We present ten examples of words with their closest words for cosine similarity in Table~\ref{tab:cosine}. We observe that our word embeddings also capture context from the sentences. For instance, the closest words to \textit{food} are mostly adjective applicable to food (e.g., \textit{solid}, \textit{average}, \textit{decent}, \textit{expensive}). We observe the same characteristic for the words of the top row in Table~\ref{tab:cosine}. For adjectives of the bottom row in Table~\ref{tab:cosine} (i.e., \textit{good}, \textit{tender}, \textit{tasty} and \textit{dry}), the closest words are either synonnyms/antonyms or nouns that may have the characteristic conveyed by the corresponding adjective. For instance, among the closest words to \textit{tender}, the words \textit{juicy} and \textit{flavorful} have similar meaning than \textit{tender}, \textit{hard} is an antonym  while \textit{gnocchi}, \textit{shrimp}, \textit{sausage} may be characterized as \textit{tender}. Finally, the closest words to \textit{time} are mostly words that convey temporal meaning (e.g., \textit{late}, \textit{day}, \textit{open}, \textit{saturday})

\begin{table*}[h]
\caption{Ten examples of cosine similarity between words (i.e., between rows of $U$) with $r=10$ on restaurant reviews dataset.}
\label{tab:cosine}
\vspace*{-0.5cm}
\begin{center}
\small
\begin{tabular}{|p{0.09\textwidth}p{0.05\textwidth}|}
\hline
food & $\mathrm{Cos}$\\
\hline
solid & 0.97 \\
delivery & 0.91 \\
average & 0.9 \\
indian & 0.9 \\
decent & 0.88 \\
overall & 0.86 \\
expensive & 0.85 \\
quality & 0.83 \\
italian & 0.83 \\
sunday & 0.82 \\
\hline
\end{tabular}
\begin{tabular}{|p{0.09\textwidth}p{0.05\textwidth}|}
\hline
service & $\mathrm{Cos}$\\
\hline
slow & 0.93 \\
friendly & 0.91 \\
fast & 0.9 \\
quick & 0.88 \\
delivery & 0.87 \\
extremely & 0.85 \\
staff & 0.83 \\
experience & 0.8 \\
average & 0.77 \\
good & 0.75 \\
\hline
\end{tabular}
\begin{tabular}{|p{0.09\textwidth}p{0.05\textwidth}|}
\hline
decor & $\mathrm{Cos}$\\
\hline
unique & 1.0 \\
vibe & 0.95 \\
warm & 0.87 \\
date & 0.87 \\
atmosphere & 0.85 \\
damn & 0.82 \\
cool & 0.81 \\
beach & 0.79 \\
broth & 0.76 \\
run & 0.73 \\
\hline
\end{tabular}
\begin{tabular}{|p{0.09\textwidth}p{0.05\textwidth}|}
\hline
atmosphere & $\mathrm{Cos}$\\
\hline
cool & 0.94 \\
unique & 0.88 \\
view & 0.85 \\
wonderful & 0.85 \\
decor & 0.85 \\
fun & 0.82 \\
vibe & 0.82 \\
date & 0.81 \\
kind & 0.79 \\
pancake & 0.78 \\
\hline
\end{tabular}
\begin{tabular}{|p{0.09\textwidth}p{0.05\textwidth}|}
\hline
meal & $\mathrm{Cos}$\\
\hline
cheap & 0.97 \\
drink & 0.97 \\
sunday & 0.96 \\
though & 0.96 \\
sushi & 0.94 \\
city & 0.93 \\
overall & 0.92 \\
visit & 0.91 \\
well & 0.91 \\
bad & 0.91 \\
\hline
\end{tabular}

\begin{tabular}{|p{0.09\textwidth}p{0.05\textwidth}|}
\hline
good & $\mathrm{Cos}$\\
\hline
location & 0.98 \\
look & 0.96 \\
hit & 0.93 \\
bad & 0.9 \\
ever & 0.9 \\
quick & 0.87 \\
okay & 0.87 \\
pretty & 0.86 \\
sure & 0.85 \\
city & 0.85 \\
\hline
\end{tabular}
\begin{tabular}{|p{0.09\textwidth}p{0.05\textwidth}|}
\hline
tender & $\mathrm{Cos}$\\
\hline
juicy & 0.96 \\
hard & 0.95 \\
flavorful & 0.93 \\
light & 0.93 \\
gnocchi & 0.89 \\
shrimp & 0.89 \\
sausage & 0.89 \\
real & 0.88 \\
water & 0.88 \\
main & 0.87 \\
\hline
\end{tabular}
\begin{tabular}{|p{0.09\textwidth}p{0.05\textwidth}|}
\hline
tasty & $\mathrm{Cos}$\\
\hline
awesome & 0.99 \\
fresh & 0.97 \\
delicious & 0.96 \\
people & 0.95 \\
course & 0.92 \\
beer & 0.92 \\
fill & 0.92 \\
fish & 0.91 \\
nice & 0.9 \\
server & 0.9 \\
\hline
\end{tabular}
\begin{tabular}{|p{0.09\textwidth}p{0.05\textwidth}|}
\hline
dry & $\mathrm{Cos}$\\
\hline
light & 0.93 \\
inside & 0.93 \\
ingredient & 0.92 \\
salty & 0.91 \\
potato & 0.9 \\
sausage & 0.89 \\
meat & 0.89 \\
put & 0.88 \\
tender & 0.85 \\
kinda & 0.85 \\
\hline
\end{tabular}
\begin{tabular}{|p{0.09\textwidth}p{0.05\textwidth}|}
\hline
time & $\mathrm{Cos}$\\
\hline
late & 0.97 \\
day & 0.97 \\
open & 0.97 \\
first & 0.96 \\
saturday & 0.93 \\
far & 0.92 \\
visit & 0.91 \\
last & 0.9 \\
though & 0.89 \\
price & 0.88 \\
\hline
\end{tabular}

\end{center}
\end{table*}

\end{document}